\def\year{2017}\relax
\documentclass[letterpaper]{article}
\usepackage{aaai17}
\usepackage{times}
\usepackage{helvet}
\usepackage{courier}
\usepackage{latexsym}
\usepackage{algorithm} %format of the algorithm
\usepackage{algorithmic} %format of the algorithm
\usepackage{multirow} %multirow for format of table
\usepackage{amsmath}
\usepackage{xcolor}
\usepackage{epsfig}
\usepackage{array}
\usepackage{bbding}
\usepackage{threeparttable}

\begin{document}
\title{A New Recurrent Neural CRF for Learning Non-linear Edge Features}
\author{Shuming Ma \and Xu Sun \\
MOE Key Laboratory of Computational Linguistics, Peking University\\
School of Electronics Engineering and Computer Science, Peking University\\
\{shumingma, xusun\}@pku.edu.cn\\
}
\maketitle
\begin{abstract}
Conditional Random Field (CRF) and recurrent neural models have achieved success in structured prediction.
More recently, there is a marriage of CRF and recurrent neural models, so that we can gain from both non-linear dense features and globally normalized CRF objective.
These recurrent neural CRF models mainly focus on encode node features in CRF undirected graphs.
However, edge features prove important to CRF in structured prediction.
In this work, we introduce a new recurrent neural CRF model, which learns non-linear edge features, and thus makes non-linear features encoded completely.
We compare our model with different neural models in well-known structured prediction tasks.
Experiments show that our model outperforms state-of-the-art methods in NP chunking, shallow parsing, Chinese word segmentation and POS tagging.
\end{abstract}

\section{Introduction}

Conditional Random Field (CRF) is a widely used algorithm for structured prediction. It is an undirected graphical model trained to maximize a conditional probability.
The undirected graph can be encoded with a set of features (node features and edge features). Usually, these features are sparse and well manual designed.

For minimizing the effort in feature engineering, neural network models are used to automatically extract features~\cite{ChenManning2014,CollobertEA2011}. These models learn dense features, which have better representation of both syntax and semantic information. Because of the success of CRF and neural networks, many models take advantage of both of them. Collobert et al.~\shortcite{CollobertEA2011} used CRF objective to compute sentence-level probability of convolutional neural networks. Durrett and Klein~\shortcite{DurrettKlein2015} introduced a neural CRF model to join sparse features and dense features for parsing. Andor et al.~\shortcite{AndorEA2016} proposed a transition-based neural model with a globally normalized CRF objective, and they use feedforward neural networks to learn neural features.

The marriage of feedforward neural network and CRF is natural because feedforward neural network scores local unstructured decisions while CRF makes global structured decisions. It is harder to combine recurrent neural model with CRF because both of them use structural inference. Huang et al.~\shortcite{HuangEA2015} provided a solution to combine recurrent structure with CRF structure, and gained good performance in sequence labelling. However, their model only encode node features while both node features and edge features are important to CRF.

In order to completely encode non-linear features for CRF, we propose a new recurrent neural CRF model. Our model uses LSTM to learn edge information of input words, and takes LSTM output as CRF energy function. We do not change the internal structure of both LSTM and CRF, so it easily decodes via standard recurrent propogation and CRF dynamic programming inference, without any extra effort. In our model, we use edge embedding to capture connections inside input structure. LSTM is used to learn hidden edge features from edge embedding. After that, CRF globally normalizes the scores of LSTM output. Andor et al.~\shortcite{AndorEA2016} proved that globally normalized CRF objective solved label bias problem for neural models.

The contribution of our paper can be listed as follow:
\begin{itemize}
  \item We propose a neural model which can learn non-linear edge features. We find that learning non-linear edge features is even more important than node features due to the ability of modelling non-linear structure dependence.
  \item We experiment our model in several well-known sequence labelling tasks, including shallow parsing, NP chunking, POS tagging and Chinese word segmentation. It shows that our model can outperform state-of-the-art methods in these tasks.
\end{itemize}

\section{Background}
\label{background}

%%%%%%%%%%%%%%%%%%%%%%%%%%%%%%%%%%%%%%%%%%%%%%%%%%
\begin{figure*}[tb]
\centering
\begin{tabular}{@{}c@{}@{}c@{}@{}c@{}@{}c@{}}

\epsfig{file=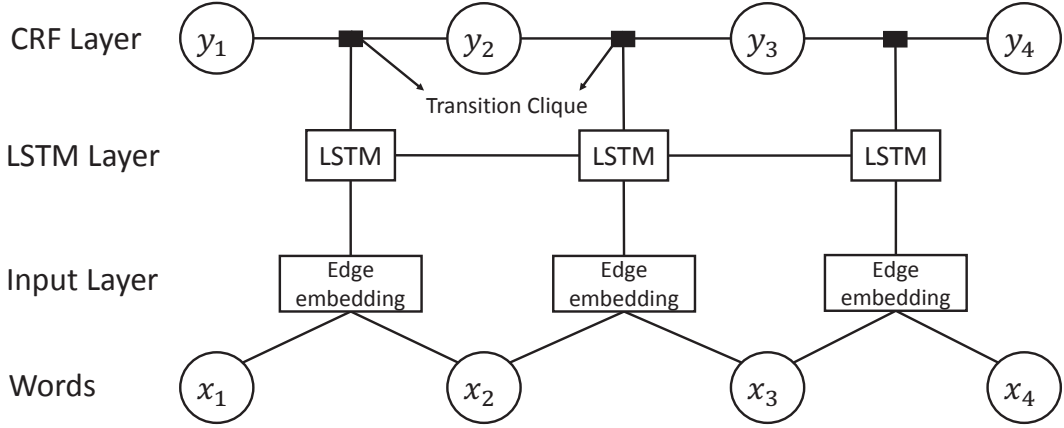,width=0.8\linewidth,clip=}

\end{tabular}
\caption{Our proposed Recurrent Neural CRF model. It learns non-linear edge features from connections of input words.
}\label{fig2}
\vspace{-0.1in}
\end{figure*}
%%%%%%%%%%%%%%%%%%%%%%%%%%%%%%%%%%%%%%%%%%%%%%%%%%

In structured prediction, our goal is to predict structure $y$ given the observations $x$. The $i^{th}$ label in structure $y$ is denoted as $y_i$, and the $i^{th}$ observation is $x_i$.

CRF~\cite{LaffertyEA2001} is a popular and effective algorithm for structured prediction. It has a log-linear conditional probability with respect to energy functions over local cliques and transition cliques:
\begin{equation}\label{crf}
\begin{split}
   \log(p(y|x)) & \propto \sum_{i}E_{local}(y_i,x,i) \\
   & +\sum_{i}E_{trans}(y_{i-1},y_i,x,i)
\end{split}
\end{equation}
where $E_{local}(y_i,x,i)$ is energy function over local clique at position $i$, and $E_{trans}(y_{i-1},y_i,x,i)$ is energy function over transition clique.

Energy functions are used to learn features. Since conventional CRF is log-linear model, both local clique and transition clique have linear energy functions:
\begin{equation}\label{local}
\begin{split}
   E_{local}(y_i,x,i) = \mu_{l}g_l(y_i,x,i))
\end{split}
\end{equation}
\begin{equation}\label{trans}
\begin{split}
   E_{trans}(y_{i-1},y_i,x,i) = \lambda_{k}f_{k}(y_{i-1},y_i,x,i)
\end{split}
\end{equation}
where $f_k$ is the indicator function of the $k^{th}$ feature for the transition clique $(y_{i-1},y_i,x)$, $g_l$ is the indicator function of $l^{th}$ feature for the local clique $(y_i,x)$, and $\lambda_{k}$ and $\mu_{l}$ are parameters of CRF.

Therefore, conventional CRF can only learn linear features. To learn high-order features, LSTM is combined with CRF model~\cite{HuangEA2015}. At each time step, LSTM recurrently inputs a word and outputs scores of each predicted labels. The output function of LSTM can be used as energy function over local cliques:
\begin{equation}\label{crfrnn1}
  E_{local}(y_i,x,i) = \sum_{i}s_i[y_i])
\end{equation}
\begin{equation}\label{crfrnn2}
  s_i = W^{(s)}h_i
\end{equation}
where $h_i$ is the hidden state of LSTM at the $i^{th}$ time step, $s_i[y_i]$ is the $y_i^{th}$ element of vector $s_i$, and $A[k,l]$ is a transition score for jumping from $j^{th}$ tag to $k^{th}$ tag.

As for transition cliques, energy function is a transition matrix of variables $A_{i,j}$ for jumping from $i^{th}$ tag to $j^{th}$ tag:
\begin{equation}\label{crfrnn3}
  E_{trans}(y_{i-1},y_i,x,i) = A_{i,j},
\end{equation}
so energy function over transition cliques is linear as conventional CRF. Therefore, LSTM-CRF learns non-linear node features (over local cliques) and linear edge features (over transition cliques).

For further contain more context information, LSTM layer can be replaced with bidirectional LSTM (BiLSTM) layer. BiLSTM contains both forward information and backward information, so that BiLSTM-CRF performs better than LSTM-CRF.

%%%%%%%%%%%%%%%%%%%%%%%%%%%%%%%%%%%%%%%%%%%%%%%%%%%%%%%%%%%%%%%%%%%%%%%%%%%%%%%%%%
\begin{table*}[tb]
\centering
\newcommand{\tabincell}[2]{\begin{tabular}{@{}#1@{}}#2\end{tabular}}
\begin{tabular}{|c|c|c|}
\hline
 & \tabincell{c}{Linear local \\ energy function} & \tabincell{c}{Non-linear local \\ energy function} \\
\hline
\tabincell{c}{Linear transition \\ energy function} & \tabincell{c}{CRF} & \tabincell{c}{LSTM-CRF \\ \cite{HuangEA2015}}\\
\hline
\tabincell{c}{Non-linear transition \\ energy function} & \tabincell{c}{\textbf{Our Edge-based-1}} & \textbf{Our Edge-based-2} \\
\hline
\end{tabular}
\caption{Different energy function of recurrent neural CRF. Our proposed models have non-linear transition energy function.} \label{table4}
\end{table*}
%%%%%%%%%%%%%%%%%%%%%%%%%%%%%%%%%%%%%%%%%%%%%%%%%%%%%%%%%%%%%%%%%%%%%%%%%%%%%%%%%%

%%%%%%%%%%%%%%%%%%%%%%%%%%%%%%%%%%%%%%%%%%%%%%%%%%%%%%%%%%%%%%%%%%%%%%%%%%%%%%%%%%
\begin{table*}[tb]
\centering
\newcommand{\tabincell}[2]{\begin{tabular}{@{}#1@{}}#2\end{tabular}}
\begin{tabular}{|c|c|c|}
\hline
 & Linear edge features & Non-linear edge features \\
\hline
\tabincell{c}{Feedforward \\ networks} & \tabincell{c}{Convolution model \\ \cite{CollobertEA2011}} & \tabincell{c}{Neural CRF networks \\ \cite{DoArtieres2010} and \\ Transition-based neural \\ networks \cite{AndorEA2016}}\\
\hline
\tabincell{c}{Recurrent \\networks} & \tabincell{c}{LSTM-CRF model \\ \cite{HuangEA2015}} & \textbf{This work} \\
\hline
\end{tabular}
\caption{Correlation between neural CRF models. Our models have recurrent structure and learn non-linear edge features.} \label{table1}
\end{table*}
%%%%%%%%%%%%%%%%%%%%%%%%%%%%%%%%%%%%%%%%%%%%%%%%%%%%%%%%%%%%%%%%%%%%%%%%%%%%%%%%%%

\section{Proposal}

Current LSTM-CRF only learns linear edge features in that it has linear energy function over transition cliques. Do and Artieres~\shortcite{DoArtieres2010} show that non-linear energy function performs better in extracting features for structured prediction. For a non-linear energy function, we propose a new recurrent neural CRF, which uses LSTM as energy function over transition cliques. Therefore, our model is able to learn non-linear edge features.

\subsection{Edge Embedding}

For learning non-linear edge features, we use edge embedding to provide raw edge information. In natural language processing, input structure is usually a sequence of words, so edges of input structure is connections of neighboring words. We have three methods to produce edge embedding from input structure.

\noindent\textbf{Bigram}: Bigram embedding is an intuition way to contain neighboring words features. We can build a bigram dictionary and assign a vector to each key. It proves to be efficient in several model~\cite{PeiEA2014,ChenEA2015}, but it may suffer from sparsity and low training speed.

\noindent\textbf{Concatenation}: Concatenation is a useful way to join two words' information. It is simple and widely used in previous work~\cite{CollobertEA2011,HuangEA2015}.

\noindent\textbf{Feedforward layer}: Feedforward layer is another method to learn information from input words. It inputs two word embedding and outputs edge embedding after a single neural network layer.

\subsection{Layers}

Figure~\ref{fig2} shows our proposed Recurrent Neural CRF model. Our model contains three layers: input layer, LSTM layer and CRF layer.

\noindent\textbf{Input Layer}: Input layer is used to input words and provide edge embedding for LSTM layer. Edge embedding is from the concatenation of neighboring word vectors, and it provides raw primary edge features.

\noindent\textbf{LSTM Layer}: LSTM layer recurrently inputs edge embedding from input layer and computes output as energy function over transition cliques for CRF layer. Our LSTM layer does not normalize energy output (using softmax function) until it does in CRF layer. Thus, our model is gobally normalized, which can solve label bias problem~\cite{AndorEA2016}.

\noindent\textbf{CRF layer}: CRF layer is to predict output structure given energy function from LSTM layer. Since we do not change CRF internal structure, viterbi algorithm is still suitable to find out the structure with highest conditional probability efficiently.

\subsection{Objective function}

In our model, objective function is similar to CRF objective, allowing computing gradients via dynamic programming. For learning non-linear features, we replace the linear energy function with LSTM output function:
\begin{equation}\label{edge1}
  p(y|x) \propto \exp(\sum_{i}t_i[y_{i-1},y_i]+\sum_{i}E_{local}(y_i,x,i))
\end{equation}
\begin{equation}\label{edge2}
  t_i = W^{(s)}h_i
\end{equation}
where $t_i[y_{i-1},y_i]$ is LSTM energy output which contains hidden edge information.

The objective has a non-linear transition energy function which neither conventional CRF nor LSTM-CRF has. The local energy function can be either linear or non-linear. We call our model with linear local energy function Edge-based-1 and the model with non-linear energy function Edge-based-2.

\noindent\textbf{Edge-based-1}: Edge-based-1 model has a linear local energy function, which captures simplest linear node features. We use it to stress the importance of learning non-linear edge features. Our experiments show that model learning only non-linear edge features outperforms model learning only non-linear node features. Local energy function in Edge-based-1 is:
\begin{equation}\label{edge3}
  E_{local}(y_i,x,i)=\mu_{l}g_l(y_i,x,i)
\end{equation}
\begin{equation}\label{edge4}
  g_l(y_i,x,i)=
  \begin{cases}
    1, & \mbox{if $y_i=l$}  \\
    0, & \mbox{otherwise}.
  \end{cases}
\end{equation}

\noindent\textbf{Edge-based-2}: Edge-based-2 model has a non-linear local energy function. It is proposed to show the combination of learning non-linear node features and edge features. Local energy function in Edge-based-2 is:
\begin{equation}\label{edge5}
  E_{local}(y_i,x,i) = \sum_{i}s_i[y_i])
\end{equation}
\begin{equation}\label{edge6}
  s_i = W^{(s)}h_i
\end{equation}
where $s_i[y_i]$ is computed by another LSTM, and contains hidden node information.

Table~\ref{table4} shows the different objective function of these recurrent nerual CRF models.

\subsection{Training}
\label{training}

We have two kinds of criteria to train our models: probabilistic criteria and large margin criteria~\cite{DoArtieres2010}.

\noindent\textbf{Probabilistic Criteria}: Probabilistic Criteria was first proposed in~\cite{LaffertyEA2001}. The regularized objective function of recurrent neural CRF can be described as:
\begin{equation}\label{objective}
  L(\theta)=\sum_{j=1}^{m}R_j(\theta)+\frac{\lambda}{2}\Vert \theta \Vert^{2}
\end{equation}
where $m$ is the number of samples in the corpus.
We denote the unnormalized score of a sample for Edge-Node Recurrent Neural CRF as:
\begin{equation}\label{score}
  F(x_j,y_j,\theta)=\sum_{i}t_i[y_{i-1},y_i] + \sum_{i}s_i[y_i]
\end{equation}
And this score for Edge-based model is:
\begin{equation}\label{score}
  F(x_j,y_j,\theta)=\sum_{i}t_i[y_{i-1},y_i] + \sum_{i}q[y_i]
\end{equation}

Then $R_j(\theta)$ in Equation~\ref{objective} can be written as:
\begin{equation}\label{score1}
  R_j(\theta)= \log\sum_{y'}\exp(F(x_j,y',\theta)) - F(x_j,y_j,\theta)
\end{equation}

\noindent\textbf{Large Margin Criteria}: Large margin criteria is first introduced by Taskar et al.~\shortcite{TaskarEA2005}. In large margin criteria, the margin between the scores of correct tag sequence and incorrect sequence will be larger than a given large margin:
\begin{equation}\label{margin1}
  F(x_j,y_j,\theta) \geq F(x_j,y',\theta) + \Delta(y_j,y')
\end{equation}
where $\Delta(y_j,y')$ is the number of incorrect tags in $y'$.

So the $R_j(\theta)$ in objective function is:
\begin{equation}\label{margin2}
  R_j(\theta)=\max_{y'}F(x_j,y',\theta)-F(x_j,y_j,\theta)+\Delta(y_j,y')
\end{equation}

\subsection{Optimization}

To minimize the objective function, we use AdaGrad~\cite{DuchiEA2011}, which is a widely used algorithm recently. The parameter $\theta_i$ for the $t^{th}$ update can be calculated as:
\begin{equation}\label{adagrad}
  \theta_{t,i} = \theta_{t-1,i}-\frac{\alpha}{\sqrt{\sum_{\tau=1}^{t}g_{\tau,i}^2}}g_{t,i}
\end{equation}
where $\alpha$ is the initial learning rate, and $g_{t,i}$ is the gradient of parameter $\theta_i$ for the $t^{th}$ update.

\section{Related Work}

Recently, neural networks models have been widely used in natural language processing~\cite{BengioEA2003,MikolovEA2010,SocherEA2013,ChenEA2015,Sun2016}. Among various neural models, recurrent neural networks~\cite{Elman1990} proves to perform well in sequence labelling tasks. LSTM~\cite{HochreiterSchmidhuber1997,GravesSchmidhuber2005} improves the performance of RNN by solving the vanishing and exploding gradient problem. Later, bidirectional recurrent model~\cite{GravesEA2013} is proposed to capture the backward information.

CRF model~\cite{LaffertyEA2001} has achieved much success in natural language processing. Many models try to combine CRF with neural networks for more structure dependence. Peng et al.~\shortcite{PengEA2009} introduces a conditional neural fields model. Collobert et al.~\shortcite{CollobertEA2011} first implements convolutional neural networks with the CRF objective.Zheng et al.~\shortcite{ZhengEA2015} integrates CRF with RNN. Durrett and Klein~\shortcite{DurrettKlein2015} uses feed forward neural networks with CRF for parsing. Huang et al.~\shortcite{HuangEA2015} use recurrent neural networks to learn non-linear node features. They show that BiLSTM-CRF is more robust than neural models without CRF. Do and Artieres~\shortcite{DoArtieres2010} suggest feedforward neural networks to learn neural features. Zhou et al.~\shortcite{ZhouEA2015} proposes a transition based neural model with CRF for parsing. Finally, Andor et al.~\shortcite{AndorEA2016} proves that a globally normalized CRF objective helps deal with label bias problem in neural models.

Compared with these neural CRF models, our recurrent neural CRF has a recurrent structure with the ability to learn non-linear edge features. Recurrent structure helps capture long distant information, and non-linear edge features provide more non-linear structure dependence. Table~\ref{table1} shows the correlation between our proposed recurrent neural CRF model and other existing neural CRF models.

%%%%%%%%%%%%%%%%%%%%%%%%%%%%%%%%%%%%%%%%%%%%%%%%%%
\begin{figure*}[tb]
\centering
\begin{tabular}{@{}c@{}@{}c@{}@{}c@{}@{}c@{}}

\epsfig{file=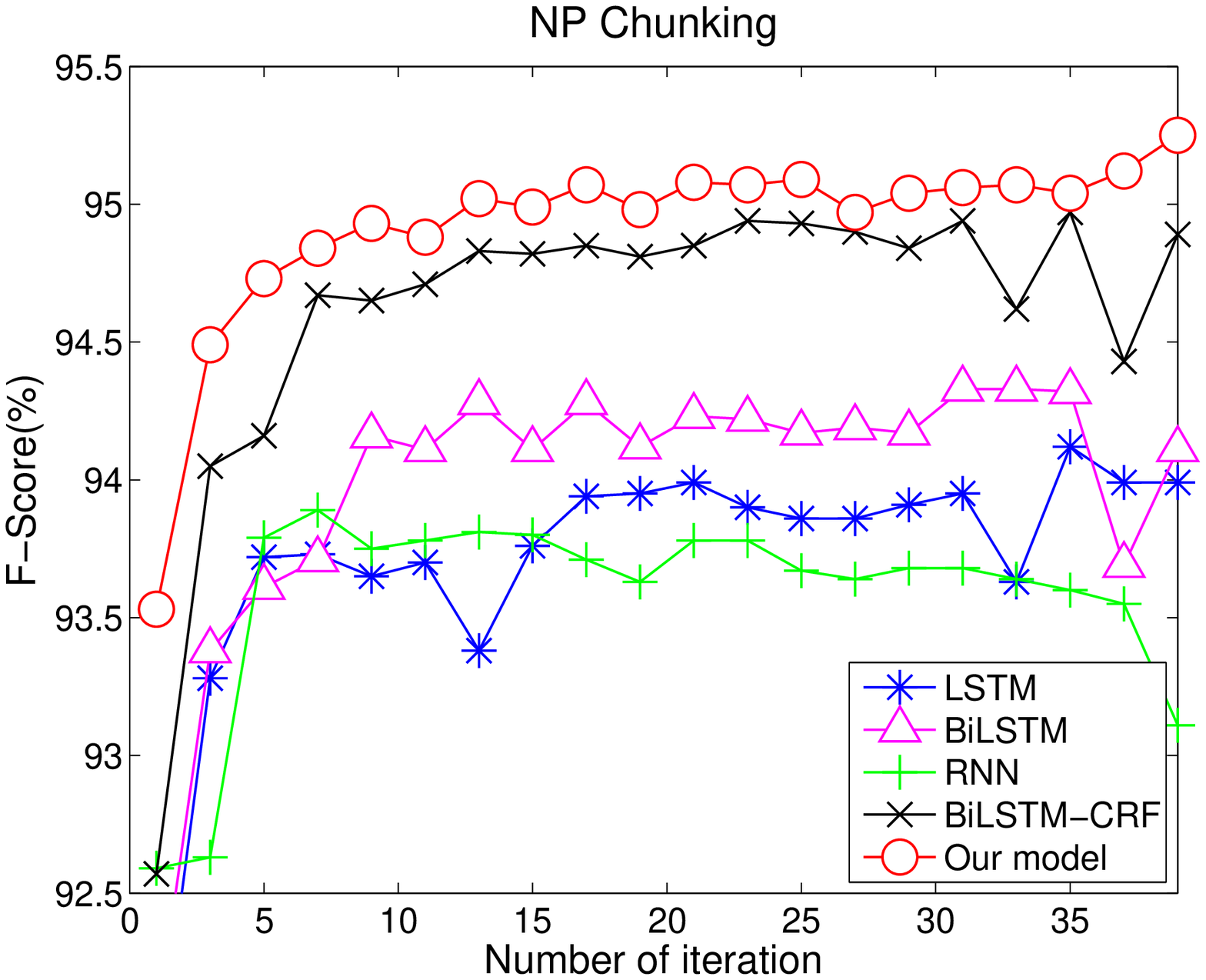,width=0.35\linewidth,clip=} &
\epsfig{file=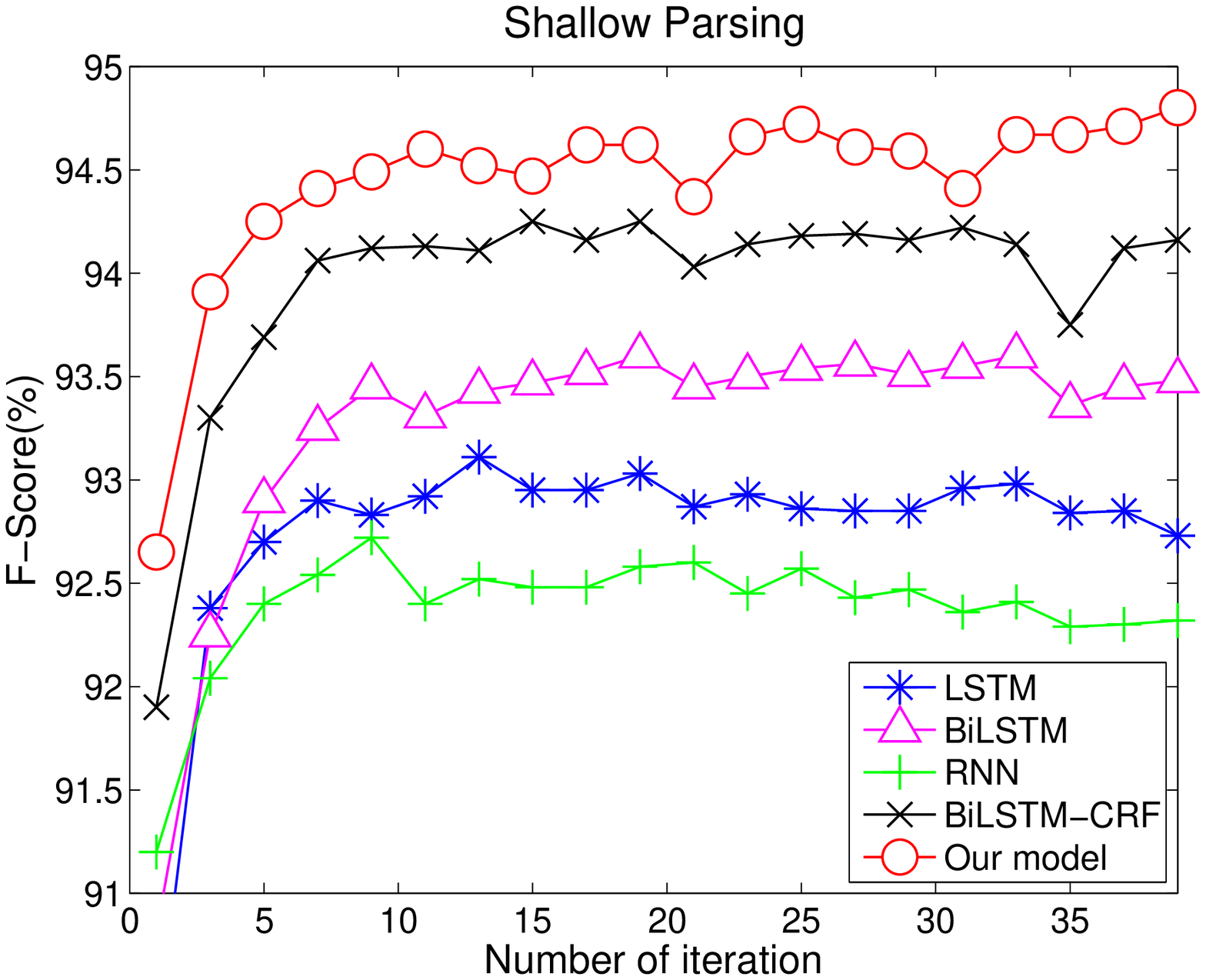,width=0.35\linewidth,clip=} \\
\epsfig{file=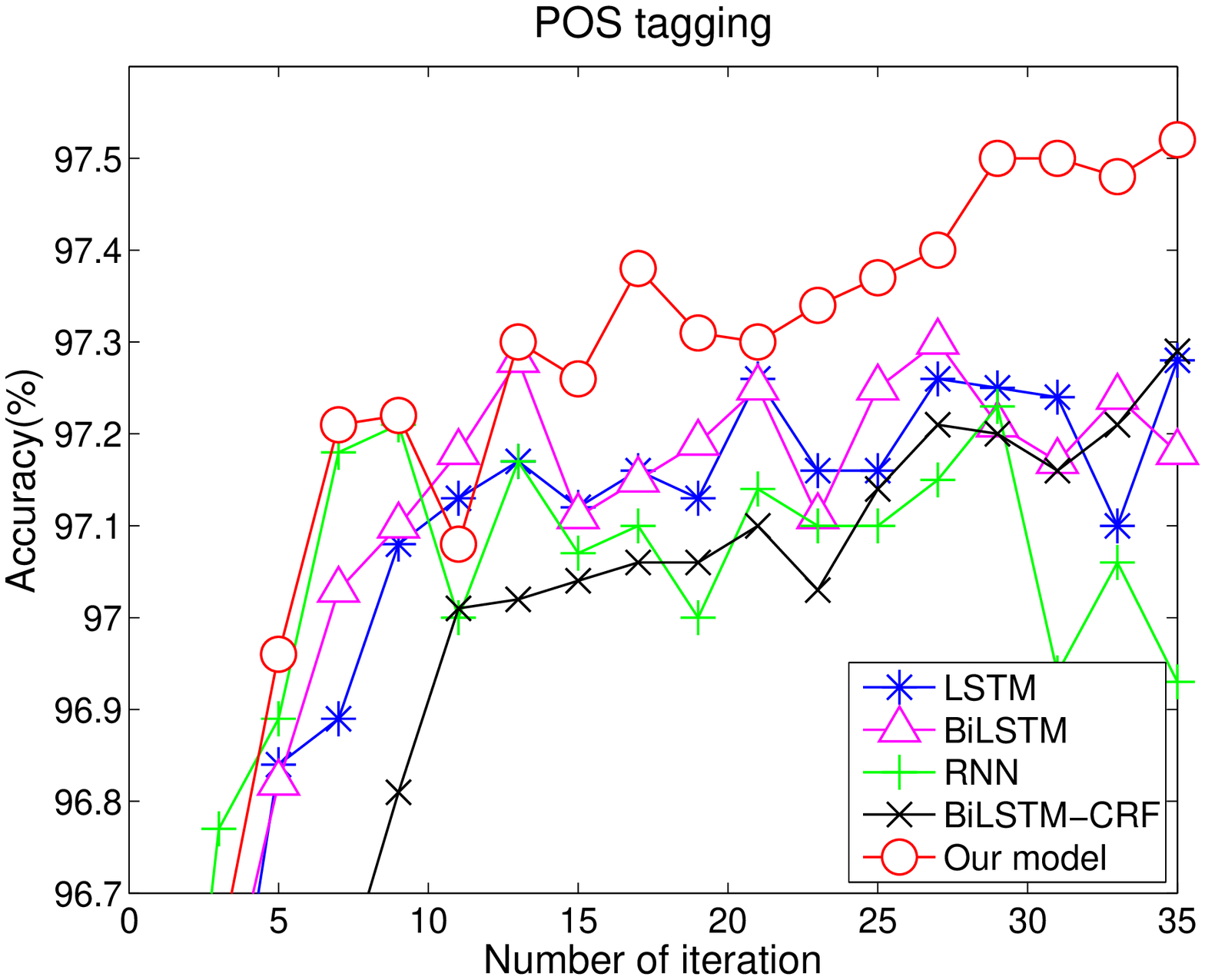,width=0.35\linewidth,clip=} &
\epsfig{file=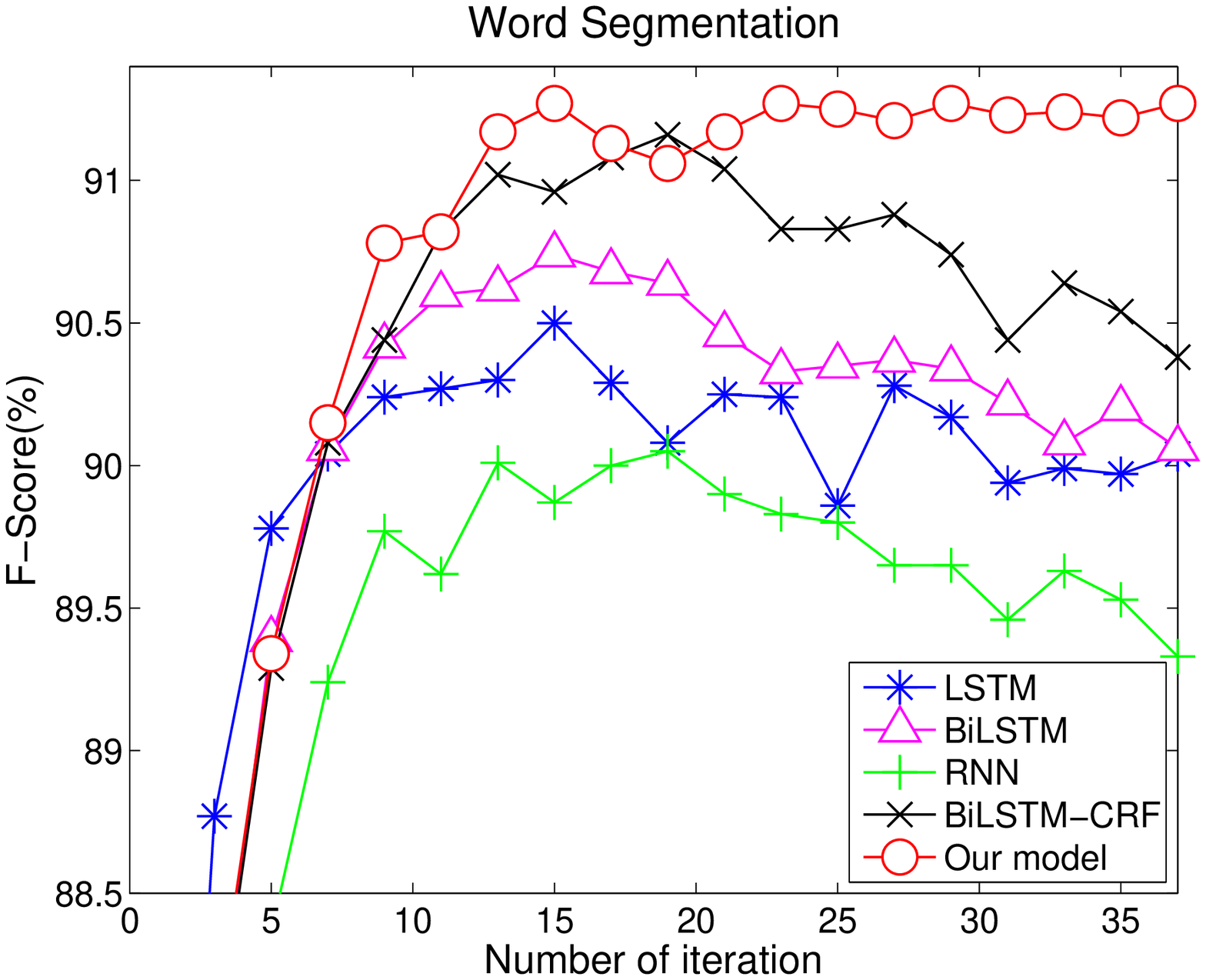,width=0.35\linewidth,clip=} \\

\end{tabular}
\caption{Performance of our Edge-based-2 model and existing recurrent neural models in test sets. It shows that our model outperforms the baseline neural models in these tasks.
}\label{fig5}
\vspace{-0.1in}
\end{figure*}
%%%%%%%%%%%%%%%%%%%%%%%%%%%%%%%%%%%%%%%%%%%%%%%%%%

%%%%%%%%%%%%%%%%%%%%%%%%%%%%%%%%%%%%%%%%%%%%%%%%%%%%%%%%%%%%%%%%%%%%%%%%%%%%%%%%%%%%%%%%%%%%%%%%%%%%%%%%%%%%%%%%%%%%%%%%%%
\begin{table*}[tb]
  \centering
  \begin{tabular}{|c|c|c|c|c|c|c|c|c|c|}
    \hline
    \multirow{2}{*}{Models} & \multirow{2}{*}{Edge} & \multicolumn{3}{c}{NP chunking} & \multicolumn{3}{|c|}{Shallow parsing} & POStag & Word Seg\\
    \cline{3-10}
    & & P & R & F1 & P & R & F1 & Acc & F1 \\
    \hline
    LSTM & $\times$ & 94.00 & 94.25 & 94.12 & 92.93 & 93.24 & 93.09 & 97.28 & 90.50 \\
    BiLSTM & $\times$ & 94.24 & 94.48 & 94.36 & 93.57 & 93.71 & 93.64 & 97.36 & 90.81 \\
    BiLSTM-CRF & linear & 94.89 & 95.05 & 94.97 & 94.33 & 94.26 & 94.29 & 97.38 & 91.16 \\
    Conv-CRF (Collobert 2011) & linear & - & - & - & - & - & 94.32 & 97.29 & - \\
    LSTM-CRF (Huang 2015) & linear & - & - & - & - & - & 94.46 & 97.55 & - \\
    \hline
    Our Edge-based-1 & non-linear & 94.86 & 95.46 & 95.16 & 94.44 & 94.52 & 94.48 & \textbf{97.56} & 91.24 \\
    Our Edge-based-2 & non-linear & \textbf{94.98} & \textbf{95.52} & \textbf{95.25} & \textbf{94.75} & \textbf{94.85} & \textbf{94.80} & 97.52 & \textbf{91.27} \\
    \hline
  \end{tabular}
  \caption{Comparison between our models and existing neural models in test sets. Our recurrent models with non-linear edge features outperform other models with linear edge features.}\label{table2}
\end{table*}
%%%%%%%%%%%%%%%%%%%%%%%%%%%%%%%%%%%%%%%%%%%%%%%%%%%%%%%%%%%%%%%%%%%%%%%%%%%%%%%%%%%%%%%%%%%%%%%%%%%%%%%%%%%%%%%%%%%%%%%%%%

%%%%%%%%%%%%%%%%%%%%%%%%%%%%%%%%%%%%%%%%%%%%%%%%%%%%%%%%%%%%%%%%%%%%%%%%%%%%%%%%%%%%%%%%%%%%%%%%%%%%
\begin{table*}[tb]
  \centering
  \begin{tabular}{|c|c|c|c|c|c|}
    \hline
    NP chunking & F1 & Shallow parsing & F1 & POS tagging & Acc \\
    \hline
    Sha and Pereira~\shortcite{ShaPereira2003} & 94.30 & Zhang et al.~\shortcite{ZhangEA2002} & 94.17 & Collobert~\shortcite{CollobertEA2011} & 97.29 \\
    Ando and Zhang~\shortcite{AndoZhang2005} & 94.70 & Ando and Zhang~\shortcite{AndoZhang2005} & 94.39 & Sun~\shortcite{Sun2014} & 97.36 \\
    Shen and Sarkar~\shortcite{ShenSarkar2005} & 95.23 & Shen and Sarkar~\shortcite{ShenSarkar2005} & 94.01 & Huang et al.~\shortcite{HuangEA2015} & 97.55 \\
    McDonald et al.~\shortcite{McDonaldEA2005} & 94.29 & Collobert et al.~\shortcite{CollobertEA2011} & 94.32 & Andor et al.~\shortcite{AndorEA2016} & 97.44 \\
    Sun et al.~\shortcite{SunEA2008} & 94.34 & Huang et al.~\shortcite{HuangEA2015} & 94.46 & Shen et al.~\shortcite{ShenEA07} & 97.33 \\
    \hline
    Our Edge-based-1 & 95.16 & Our Edge-based-1 & 94.48 & Our Edge-based-1 & \textbf{97.56} \\
    Our Edge-based-2 & \textbf{95.25} & Our Edge-based-2 & \textbf{94.80} & Our Edge-based-2 & 97.52 \\
    \hline
  \end{tabular}
  \caption{Comparison of our models with recent methods for NP Chunking, shallow parsing and POS tagging in test sets. Our models outperform state-of-the-art methods in these tasks. Our corpus for Chinese word segmentation in social media text is so latest that we do not find comparable result.}\label{table3}
\end{table*}
%%%%%%%%%%%%%%%%%%%%%%%%%%%%%%%%%%%%%%%%%%%%%%%%%%%%%%%%%%%%%%%%%%%%%%%%%%%%%%%%%%%%%%%%%%%%%%%%%%

\section{Experiments}

We perform some experiments to analyze our proposed models. We choose well-known sequence labelling tasks, including NP chunking, shallow parsing, POS tagging and Chinese word segmentation as our benchmark so that our experiment results are comparable. We compare our model with other popular neural models, and analyze the effect of non-linear edge features.

\subsection{Tasks}

We introduce our benchmark tasks as follows:

\noindent\textbf{NP Chunking}: NP Chunking is short for Noun Phrase Chunking, that the non-recursive cores of noun phrases called based NPs are identified. Our datasets are from \emph{CoNLL-2000 shallow-parsing shared task}, which consists of 8936 sentences in training set and 2012 sentences in test set. We further split the training set and extract 90\% sentences as development set. Following previous work, we label the sentences with BIO2 format, including 3 tags (B-NP,I-NP,O). Our evaluation metric is F-score.

\noindent\textbf{Shallow Parsing}: Shallow parsing is a task similar to NP Chunking, but it needs to identify all chunk types(VP,PP,DT...). The dataset is also from \emph{CoNLL-2000}, and it contains 23 tags. We use F-score as the evaluation metric.

\noindent\textbf{POS tagging}: POS tagging is short for Part-of-Speech Tagging, that each word is annotated with a particular part-of-speech. We use the standard benchmark dataset from the Penn Treebank. We use Sections 0-18 of the treebank as the training set, Sections 19-21 as the development set, and Sections 22-24 as the test set. We use tag accuracy as evaluation metric.

\noindent\textbf{Chinese word segmentation for social media text}: Word segmentation is a fundamental task for Chinese language processing~\cite{SunEA2012,XuSun2016}. Although current models perform well in formal text, many of them do badly in informal text like social media text. Our corpus is from NLPCC2016 shared task. Since we have no access to test set, we split training set and extract 10\% samples as test set. We use F-score as our evaluation metric.

%\subsection{Criteria Selection}

%We introduce two kinds of criteria for our model. Probabilistic criteria uses a traditional CRF loss function, which needs forward-backward algorithms to compute the gradients. Large margin criteria uses a margin based loss function, which needs viterbi algorithm to work out the sequence with the maximum score. Therefore, probabilistic criteria will lead to a more steady convergence but it costs more time in computing the gradient. Large margin criteria saves much time thanks to viterbi algorithm but it may lead to some fluctuation of performance during training. In practice, we compare two probabilistic criteria and select probabilistic criteria in our model.

\subsection{Embeddings}

Embeddings are distributed vectors to represent the semantic of words~\cite{BengioEA2003,MikolovEA2013}. It proves that embeddings can influence the performance of neural models. In our models, we use random initialized word embeddings as well as Senna embeddings~\cite{CollobertEA2011}. Our experiments show that Senna Embeddings can slightly improve the performance of our models. We also incorporate the feature embeddings as suggested by previous work~\cite{CollobertEA2011}. The features include a window of last 2 words and next 2 words, as well as the word suffixes up to 2 characters. Besides, we make use of part-of-speech tags in NP chunking and shallow parsing. To alleviate heavy feature engineering, we do not use other features like bigram or trigram, though they may increase the accuracy as shown in ~\cite{PeiEA2014} and ~\cite{ChenEA2015}. All these feature embeddings are random initialized.

We also try three methods to learn edge embedding, including concatenate current words embeddings with feature embeddings as our edge embedding in our model.

\subsection{Settings}

We tune our hyper-parameters on the development sets. Our model is not sensitive to the dimension of hidden states when it is large enough. For the balance of accuracy and time cost, we set this number to 300 for NP chunking and shallow parsing, and the number is 200 for POS tagging and Chinese word segmentation. The dimension of input embeddings is set to be 100. The initial learning rate of AdaGrad algorithm is 0.1, and the regularization parameter is $10^{-6}$. The dropout method proves to avoid overfitting in neural models~\cite{SrivastavaEA2014}, but we find it has limited impact in our models. Besides, we select probabilistic criteria to train our model for its steady convergence and robust performance.

\subsection{Baselines}

We choose current popular neural models as our baselines, including RNN, LSTM, BiLSTM and BiLSTM-CRF. RNN and LSTM are basic recurrent neural models. For further learn bidirectional context information, we also implement Bi-LSTM for our tasks. We compare our model with these model to show the gain from combining neural model with CRF objective. Finally, BiLSTM-CRF is our strong baseline. We compare our model with BiLSTM-CRF to show that learning non-linear edge features is more important than single non-linear node features.

\subsection{Results}

%%%%%%%%%%%%%%%%%%%%%%%%%%%%%%%%%%%%%%%%%%%%%%%%%%
%\begin{figure*}[tb]
%\centering
%\begin{tabular}{@{}c@{}@{}c@{}@{}c@{}@{}c@{}}

%\epsfig{file=pic/np_2.eps,width=0.33\linewidth,clip=} &
%\epsfig{file=pic/text_2.eps,width=0.33\linewidth,clip=} &
%\epsfig{file=pic/pos_2.eps,width=0.33\linewidth,clip=} \\

%\end{tabular}
%\caption{Performances of two criteria for our Edge-based models in test sets. It shows that probabilistic criteria has much better performance in NP chunking and shallow parsing, as well as slightly higher accuracy in POS tagging.
%}\label{fig5}
%\vspace{-0.1in}
%\end{figure*}
%%%%%%%%%%%%%%%%%%%%%%%%%%%%%%%%%%%%%%%%%%%%%%%%%%

We analyze the performance of our models in the above benchmark tasks. Our baselines include popular neural models. We train each model for 40 passes through the training sets. The performance curves of these models in test sets are provided as showed in Figure~\ref{fig5}. It shows that our Edge-based model outperforms the baseline neural models, including RNN, LSTM, BiLSTM and BiLSTM-CRF.

According to Table~\ref{table2}, our models significantly outperform recurrent models without edge information in three tasks. It concludes that globally normalized objective can bring better performance in that it can model more structure dependence. Besides, our models also have higher accuracy than models with linear edge features, which shows that modelling non-linear edge features is very important for neural models. It seems that Edge-based-2 achieves better result than Edge-based-1 in NP chunking and shallow parsing, so combining non-linear edge features with node features is helpful in these two tasks.

We also compare our models with some existing systems as shown in Table~\ref{table3}.

\noindent\textbf{NP Chunking}: In NP Chunking, a popular algorithm is second-order CRF~\cite{ShaPereira2003}, which can achieve a score of 94.30\%. McDonald et al.~\shortcite{McDonaldEA2005} implemented a multilabel learning algorithm, with a score of 94.29\%. Sun et al.~\shortcite{SunEA2008} proposed a latent variable CRF model, improving the score up to 94.34\%. Some other models~\cite{AndoZhang2005,ShenSarkar2005} make use of extra resources, and greatly improve the performance of Support Vector Machines(SVM). To the best of our knowledge, few neural models have been introduced for NP Chunking. Our models can outperform all of the above models. We also implement some neural models to compare with our model. LSTM has a score of 94.12, and BiLSTM is better with 94.36\% F-score. As a strong baseline, BiLSTM-CRF outperforms them with 94.97\% F-score. Our model also performs better than all these neural models, with 95.25\% F-score.

\noindent\textbf{Shallow Parsing}: In shallow parsing, Zhang et al.~\shortcite{ZhangEA2002} proposed a generalized Winnow algorithm which achieve a score of 94.17\%. Ando and Zhang~\shortcite{AndoZhang2005} introduced a SVD based alternating structure optimization algorithm, improving the score up to 94.39\%. Collobert et al.~\shortcite{CollobertEA2011} first introduced the neural network model to shallow parsing. They combined the convolutional neural networks with CRF, and reached 94.32\% F-score. Huang et al.~\shortcite{HuangEA2015} combined BiLSTM with a CRF layer, raising the score up to 94.46\%. Our Edge-based model can beat all of these models in performance, and obtain state-of-art result with a score of 94.80\%.

\noindent\textbf{POS tagging}: As an important task in natural language processing, there are lots of work on POS tagging. We make a comparison of our models with some recent work. Sun~\shortcite{Sun2014} introduced a structure regularization method for CRF, which reached 97.36\% accuracy. Collobert et al.~\shortcite{CollobertEA2011} used a Convolution-CRF model, and obtained 97.29\%. Andor et al.~\shortcite{AndorEA2016} proposed a globally normalized transition based neural model, which made use of feedforward neural networks and achieved 97.44\% accuracy. Our Edge-based model can outperform the above models with 97.56\% accuracy.

\noindent\textbf{Chinese word segmentaion for social media text}: Our corpus is latest so we do not find comparable result. Instead, we implement some state-of-the-art models, and compare with our model. We find that LSTM achieves 90.50\% F-score while BiLSTM is slightly better with 90.81\%. BiLSTM gains from CRF objective, and achieves 91.16\% F-score. Our model can beat all of these model, with a 91.27\% F-score.

\subsection{Significance Tests}

We conduct significance tests based on t-test to show the improvement of our models over the baselines. The significance tests suggest that our Edge-based-1 model has a very significant improvement over baseline, with $p \leq 0.004$ in NP chunking, $p \leq 0.007$ in shallow parsing and $p \leq 0.0001$ in POS tagging and Chinese word segmentation. The Edge-based-2 model also has high statistically significance, with $p \leq 0.0001$ in all tasks. The significance tests support theoretical analysis that our models can outperform the baselines in accuracy.

\section{Conclusions}

We propose a new recurrent neural CRF model for learning non-linear edge features. Our model is capable to completely encoding non-linear features for CRF.
Experiments show that our model outperforms state-of-the-art methods in several structured prediction tasks, including NP chunking, shallow parsing, Chinese word segmentation and POS tagging.

\section{Acknowledgements}

This work was supported in part by National Natural Science Foundation of China (No. 61300063). Xu Sun is the corresponding author of this paper.

\bibliographystyle{aaai}
\bibliography{aaai2017}

\end{document}